\documentclass[10pt,twocolumn,letterpaper]{article}
\pdfoutput=1
\usepackage{wacv}
\usepackage{times}
\usepackage{epsfig}
\usepackage{graphicx}
\usepackage{amsmath}
\usepackage{amssymb}
\usepackage{multirow}

\newcommand{\mb}{\mathbf}

\usepackage[pagebackref=true,breaklinks=true,letterpaper=true,colorlinks,bookmarks=false]{hyperref}

\wacvfinalcopy 

\def\wacvPaperID{906} 

\ifwacvfinal\fi
\begin{document}

\title{Action Graphs: Weakly-supervised Action Localization\\with Graph Convolution Networks}

\author{Maheen Rashid\\
University of California, Davis\\
{\tt\small mhnrashid@ucdavis.edu}
\and
Hedvig Kjellstr\"om\\
KTH Royal Institute\\
of Technology\\
{\tt\small hedvig@kth.se}
\and
Yong Jae Lee\\
University of California, Davis\\
{\tt\small yongjaelee@ucdavis.edu}
}

\maketitle

\begin{abstract}
\vspace{-10pt}
We present a method for weakly-supervised action localization based on graph convolutions. In order to find and classify video time segments that correspond to relevant action classes, a system must be able to both identify discriminative time segments in each video, and identify the full extent of each action. Achieving this with weak video level labels requires the system to use similarity and dissimilarity between moments across videos in the training data to understand both how an action appears, as well as the sub-actions that comprise the action's full extent. However, current methods do not make explicit use of similarity between video moments to inform the localization and classification predictions. We present a novel method that uses graph convolutions to explicitly model similarity between video moments. Our method utilizes similarity graphs that encode appearance and motion, and pushes the state of the art on THUMOS’14, ActivityNet 1.2, and Charades for weakly-supervised action localization. \end{abstract}

\vspace{-0.1in}
\section{Introduction}
\begin{figure}[t]
    \centering
    \includegraphics[width=\linewidth]{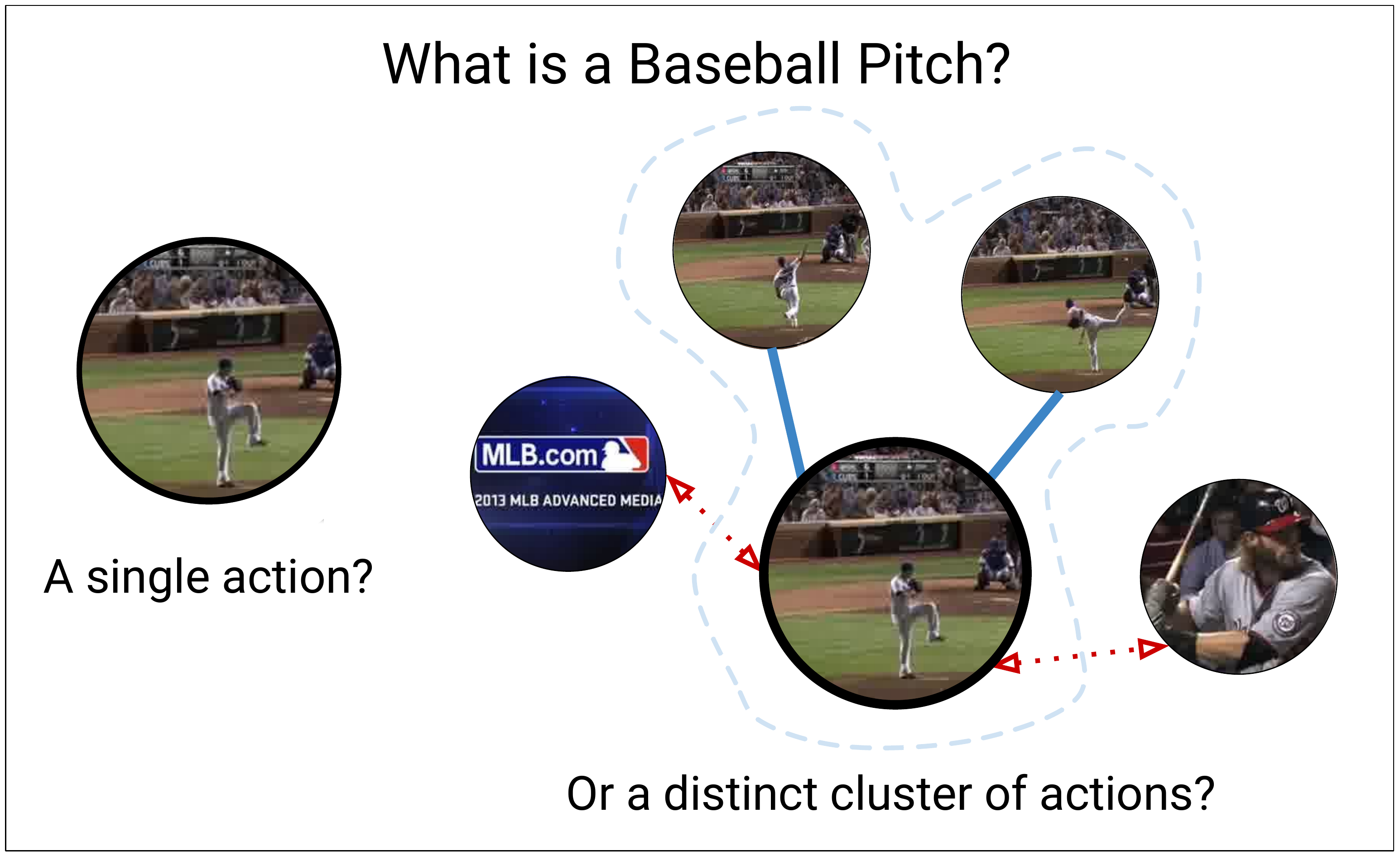}
    \caption{\textbf{Key Idea:} A baseball pitch is not defined by a single action -- rather it is defined by a series of smaller actions that are distinct from other actions in a video. Despite this, prior methods classify every time segment individually before collating predictions for localization (left). We instead explicitly model what each segment is similar to -- blue edges -- and different from -- red edges -- for weakly-supervised temporal action localization (right).}
    \label{fig_intro}
\end{figure}

Temporal activity localization is the problem of identifying the start and end times of every action's occurrence \cite{aggarwal2011human,herath2017going}. In a fully supervised setting, every training video is annotated with the start and end time of each action's occurrence. However, acquiring manual temporal annotations is an onerous task and severely limits both the number and diversity of actions that a system can be trained to identify. In contrast, systems that can successfully classify and temporally localize actions with \emph{weak} video-level labels---that only state whether an activity is present in the video or not---provide a more scalable solution.

Without frame-level annotations, weakly-supervised systems must rely on similarity cues between video time segments. Specifically, they must (1) use the dissimilarity between foreground segments of different action classes to classify videos correctly; (2) use the similarity/relationship between foreground segments of the same action to determine its full extent; and (3) infer that the similarity between segments of different actions' videos are indicative of background segments.

Although great progress has been made on this challenging problem, existing approaches~\cite{yuan2018marginalized,Paul_2018_ECCV,nguyen2017weakly,wang2017untrimmednets} do not explicitly model the \emph{relationships between time segments} to inform their final predictions. Instead, most approaches first split the video into multiple time segments, and classify each segment separately. These segment-level predictions are then pooled to perform the final video-level classification using multiple instance learning~\cite{zhou2004multi}. The relationships between time segments are either only implicitly used during training to learn attention~\cite{nguyen2017weakly}, perform the final video-level classification~\cite{wang2017untrimmednets}, or to create good features~\cite{yuan2018marginalized,Paul_2018_ECCV}, but are not used during test time. In contrast, Xu et al.~\cite{xu2018segregated} use a recurrent neural network to model relationships between time segments. However, similarity between time segments that are temporally distant, or belong to different videos cannot be modeled in their framework. In other words, the model lacks the ability to ensure that all time segments regardless of temporal location that are related to the same action are treated similarly. 

\vspace{-10pt}
\paragraph{Main idea.} Our main idea is to explicitly model the similarity relationships between time segments of videos in order to classify and localize actions in videos. We use graph convolution networks (GCNs) \cite{thomas2016semi} for this purpose. 

Similar to regular convolution networks, GCNs also perform nonlinear transformations on the input features. However, in addition, GCNs treat input features as nodes in a graph with weighted edges. By setting the edge weights to be proportional to the level of similarity between nodes, GCNs allow feature similarity and dissimilarity to be incorporated into the weight learning process as gradients are propagated across weighted edges, as well as during test time as inference is performed over an entire graph.

By using GCNs, our method explicitly ensures that relationships between time segments are considered during both training and testing. We represent each segment in a video as a node in a graph, and edges between nodes are weighted by their similarity. Each segment's feature representation is transformed to a weighted average of all segments it is connected to, with weights based on learned edge strength. These weighted average features are then used to learn a multiple instance learning based video classifier. We use appearance and motion similarity between segments to determine edge weights: two nodes that have similar RGB and optical flow features have a stronger edge between them than two nodes that have dissimilar RGB and optical flow features. In this way, the learned weights operate on groups of features together, rather than on individual time segments. This helps prevent the network from focusing on just a  few discriminative parts of the video. 

\vspace{-10pt}
\paragraph{Contributions.} (1) A novel graph convolution approach for weakly-supervised action localization. Our method is based on an appearance and motion similarity graph and is the first to use graph convolutions in the weakly-supervised action localization setting. (2) We analyze each component of our model, explore other graph based alternatives, and quantitatively and qualitatively compare against other non-graph based approaches. 
(3) We push the state-of-the-art on widely-used action detection datasets in the weakly-supervised setting - THUMOS’14 \cite{idrees2017thumos} and ActivityNet 1.2 \cite{caba2015activitynet}, and are the first to present results on Charades~\cite{sigurdsson2016hollywood}. 

\vspace{-0.05in}
\section{Related work}

Weakly supervised action localization has many different variants in literature. \cite{Paul_2018_ECCV} encourages time segments with similar classification predictions to have similar intermediate deep features using a Co-Activity Similarity Loss. Like us, it uses feature similarity between segments to improve localization. However, unlike our approach, it exclusively uses feature similarity to provide training supervision, and does not model feature relationships to make predictions. Others discourage the network from focusing only on the most discriminative time segments via random hiding~\cite{singh2017hide}, or their iterative removal during training~\cite{zeng2019breaking}. While ~\cite{Shou_2018_ECCV} uses a contrastive loss for temporally fuller localizations, ~\cite{zhai2019action} additionally uses a coherence loss for visually consistent action identification. More recent works learn to attend and pool per time segment predictions during training~\cite{nguyen2017weakly,yuan2018marginalized}, while UntrimmedNets~\cite{wang2017untrimmednets} simultaneously learns to classify and select the most salient segments in a video. However, these methods do not consider the relationships between time segments during testing. In contrast, by inferring over a video-level graph, our method can use information from the entire video during training \emph{as well as testing} to achieve better localization. Recent work \cite{xu2018segregated} uses recurrent neural networks to model relationships between time segments. However, relationships between time segments that are temporally distant, or that belong to different videos cannot be modeled. In contrast, our model is not restricted by temporal proximity when modeling similarity and dissimilarity relationships between time segments.

Some work use additional cues such as person detection~\cite{siva2011weakly,weinzaepfel2016towards}, scripts/subtitles~\cite{laptev2008learning,duchenne2009automatic,bojanowski2013finding}, or external text~\cite{richard2018action}.  Others use activity ordering information to assist in discriminative clustering \cite{bojanowski2014weakly,bojanowski2015weakly}, temporal alignment~\cite{huang2016connectionist,kuehne2017weakly,richard2017weakly}, and segmenting temporal proposals~\cite{Farha_2018_CVPR,richard2018viterbi}.

A growing body of work explore neural network based graphs \cite{thomas2016semi,battaglia2018relational}. In computer vision, graph convolutions have gained popularity for capturing relationships between objects spatially and temporally for video object understanding, as well as capturing spatio-temporal dynamics for action understanding~\cite{yuan2017temporal,wang2018non,Wang_2018_ECCV,ghosh2018stacked,chen2018graphbased,zhang2018structured,huang2018dynamic}. In particular, \cite{yuan2017temporal} develops an LSTM based graph for video object detection that uses strong action localization annotation as supervision. Unlike our method they do not use graph convolutions, and operate in a different `slightly supervised' setting for video object detection, where human action labels are used to generate object detection labels. \cite{Wang_2018_ECCV} uses both an appearance similarity graph alongside a temporal similarity graph to understand relations between video regions for action classification. However, unlike our method, it operates in a fully-supervised setting.

\vspace{-0.05in}
\section{Approach}

\begin{figure*}[t]
    \centering
    \includegraphics[width=0.95\textwidth]{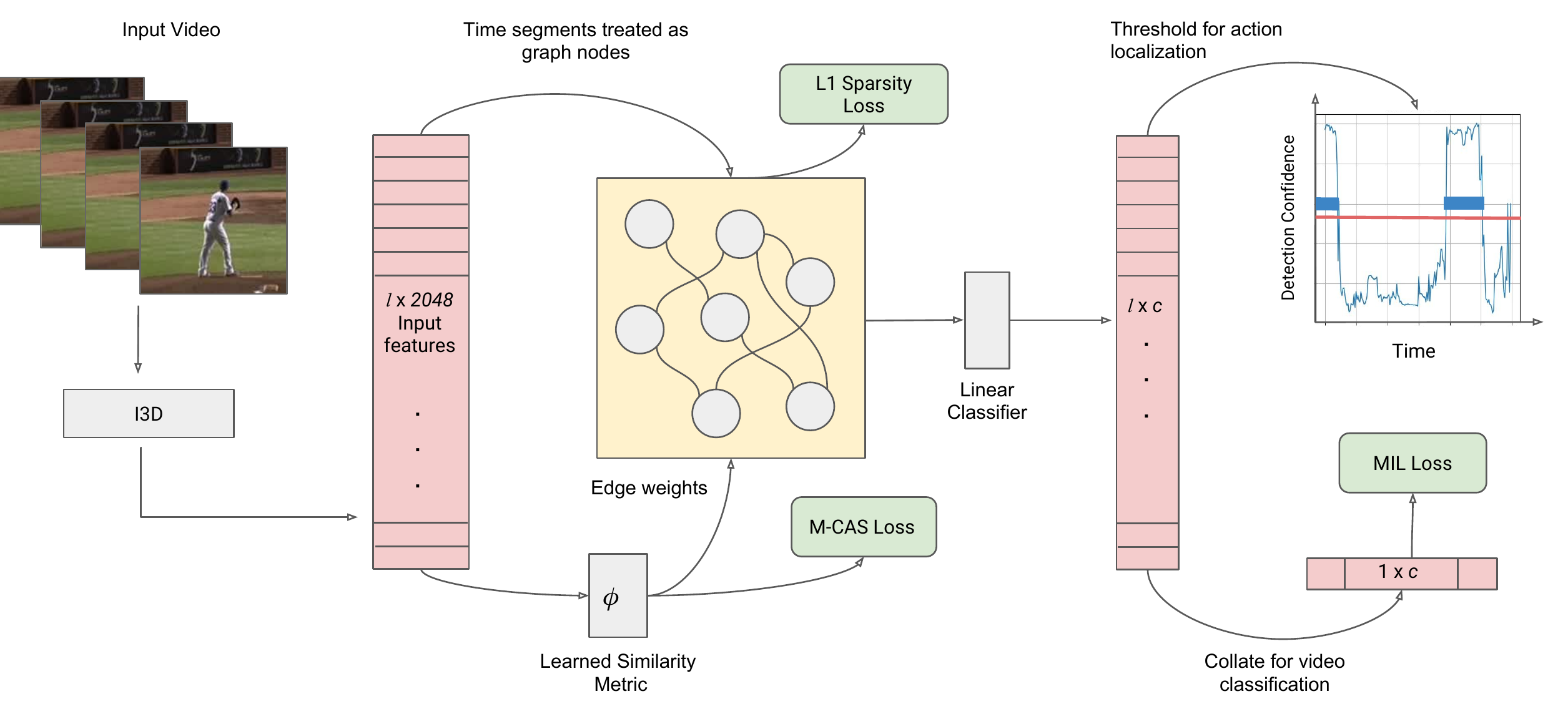}
    \caption{\textbf{Method Overview:} We use a pre-trained I3D network to extract input features for each time segment in a video. Each time segment is represented as a graph node, and edges between nodes are weighted by their level of learned similarity. Segment-level classification predictions are made by inference over this graph. During test time, we threshold the segment-level predictions to get activity localization predictions. We use a Multiple-Instance Learning (MIL) loss to supervise the classification, an L1 loss on edge weights to keep the edges in the graph sparse, and a modified Co-Activity Similarity Loss (M-CASL) to encourage edges between foreground segments to be higher than edges between foreground and background segments.}
    \vspace{-0.1in}
    \label{fig_approach}
\end{figure*}

Our goal is to train a temporal action localization system that predicts the start and end times of each action's occurrence in a video. During training, we are only provided with weak action labels: we know what actions occur in a video but we do not know when or how many times they occur. We use these weak action label--video pairs to train our system. During testing, input videos have no labels.

\subsection{Architecture}
Our network architecture is shown in Fig.~\ref{fig_approach}. The input to our network is an $l \times d_{in}$ volume of features, where $l$ is the number of input time segments in the video, and $d_{in}$ is feature dimension. We refer to each time segment's input feature as $\mb{x}$ and the entire input volume as $\mb{X}$. The input features are then transformed using a graph convolution layer.  We use RGB and optical flow based similarity to weight edges in the graph, where the similarity metric is learned by a separate linear layer $\phi$. For each input time segment, the network outputs a prediction confidence for all classes.  We refer to the final prediction $l\times c$ volume as $\mb{Y}$, where $c$ is the number of action classes. 

\subsection{Feature extraction}
We extract features from a Kinetics pre-trained I3D~\cite{carreira2017quo} to represent each video segment, as in \cite{Paul_2018_ECCV}. Specifically, each video is represented by two $l\times1024$ volumes (where $l$ is the number of input time segments), one extracted from a RGB based stream and one extracted from an optical flow based stream. These volumes are concatenated to give a final $l\times2048$ representation. Each time segment corresponds to 16 frames extracted at 25 FPS, or 0.64 seconds.

\subsection{Graph convolution layer}\label{sec:graphconvlayer}

Each input time segment is treated as a node in a graph over which inference is performed. The node edges are weighted by their similarity. In this way, related time segments can be pushed together and unrelated time segments can be pushed apart in feature space, while informing one another during both training and testing phases. Through this process, the graph convolutions can encourage better localization as the network is forced to inspect and predict each time segment class in the context of other time segments that it is similar to as well as different from.

The graph layer performs the following transformation on input $\mb{X}$:
\[
	\mb{Z} = \mb{\widehat{G}XW}
\]
where $\mb{Z}$ is an $l \times d_{out}$ output of the graph convolution, $\mb{W}$ is a $2048 \times d_{out}$ weight matrix learned via backpropagation, and $\mb{\widehat{G}}$ is the row normalized affinity matrix $\mb{G}$. $\mb{G}$ is an $l \times l$ affinity matrix where $\mb{G}_{ij}$ is the edge weight between $\mb{x}_i$ and $\mb{x}_j$.

To compute $\mb{G}$, we first learn a simple affine function $\phi$ on input feature $\mb{x}$:
\[
	\phi(\mb{x}) = \mb{wx} + \mb{b}
\]
where $\mb{w}$ and $\mb{b}$ are weight and bias terms.  $\phi$ is used to weight graph edges such that nodes with more similar $\phi$ have higher edge weights between them. $\mb{G}_{ij}$ (edge weight between $\mb{x_i}$ and $\mb{x_j}$) is computed as:
\begin{equation*}\label{eq_edge_sim}
\mb{G}_{ij} = f(\phi(\mb{x}_i),\phi(\mb{x}_j))
\end{equation*}
where $f(\cdot)$ is cosine similarity.

$\mb{G}$ essentially transforms each row of $\mb{X}$ to a weighted combination of other rows of $\mb{X}$. Note that this formulation subsumes other common layer operations. An identity $\mb{G}$ corresponds a regular fully connected layer with no bias term. A $\mb{G}$ with zero off-diagonal values, and non uniform diagonal entries works similarly to an attention mechanism. By setting rows of $\mb{G}$ to one or zero, average and max pooling operations can be performed. Multiple graph layers can be stacked together as the $\mb{Z}$ of the layer below becomes the $\mb{X}$ of the layer above. However, due to the small size of our datasets we use only a single graph layer. The output of our graph layer is passed to a linear classification layer to obtain the final $l \times c$ volume $\mb{Y}$.

\subsection{Loss functions}
Our method uses three separate losses. We use a multi-instance cross entropy loss that trains the network to correctly classify each video via segment level classification. We also impose an L1 sparsity loss on our graph so that graph edges are sparse and discriminative time segments can be clustered together. Last, we impose a co-activity similarity loss on the learned similarity function $\phi$, so that salient parts for each video class are encouraged to have high edge weights between them.

\subsubsection{Multiple instance learning loss}\label{sec_mil}
Similar to prior work~\cite{wang2017untrimmednets,Paul_2018_ECCV}, we treat the problem of weak action localization as a multiple instance learning (MIL) problem. Each video is treated as a bag of instances, some of which are positive instances. We only have video-level labels, and must use them to correctly classify instances within each video. To do this, we classify all instances, and then average the classification predictions for the top $k$ per class to get a $c$ dimension video-level prediction vector.
The vector is normalized using softmax so that each dimension, $p_i$ represents the probability for class $i$. At the same time, the binary indicator ground truth vector $y$ (a video can contain multiple action classes) for a video is normalized so that it sums to $1$. It is then used alongside the video prediction vector to calculate the multi-class cross entropy loss averaged across a batch of $n$ videos, indexed by $j$:
\[
L_{MIL} = \frac{1}{n} \sum_{j=1}^{n} \sum_{i=1}^{C} - y_i^j \log p_i^j.
\]

We set $k$ to $\max (1,\lfloor \frac{l}{d} \rfloor)$ where $l$ is the total number of input features for a video, and $d$ is a hyper parameter. We further analyze the effect of $d$ in Section \ref{sec:denosize}. This part of our framework is similar to the multiple instance learning loss branch in~\cite{Paul_2018_ECCV} and the hard selection module of~\cite{wang2017untrimmednets}.  Unlike binary cross entropy loss, this loss formulation gives equal weight to each training video rather than each label occurrence. Hence, instances of each class that occur in videos with fewer labels get more weight than instances that co-occur with many other classes, which we found leads to better performance than the binary cross entropy loss.

\subsubsection{Graph sparsity loss}

To recap, $\mb{G}$ transforms rows of $\mb{X}$ to a weighted average of rows of $\mb{X}$. In other words, $\mb{G}$ can cluster together similar $\mb{x}$'s, and push apart dissimilar $\mb{x}$'s. However, a $\mb{G}$ with edge weights that are close to uniform will make it hard for the network to train, as discriminative signals in $\mb{X}$ will be averaged out. In order to prevent this, we enforce edge weights in $\mb{G}$ to be sparse by imposing an L1 loss on the absolute sum of $\mb{G}$:

\begin{equation*}
    L_{L1} = \frac{\sum_{i=1}^{l}\sum_{j=1}^{l} | \mb{G}_{ij} |}{l^2}
\end{equation*}
The loss works to encourage sparsity in $\mb{G}$, and hence trains $\phi$ to create tighter clusters from $\mb{X}$.

We find that it is helpful to additionally ignore edges that have a low absolute value. We therefore drop edges in each graph that are in the lower half of its range of edge weights.

\subsubsection{Modified co-activity similarity loss}
\begin{figure*}[t]
    \centering
    \includegraphics[width=\textwidth]{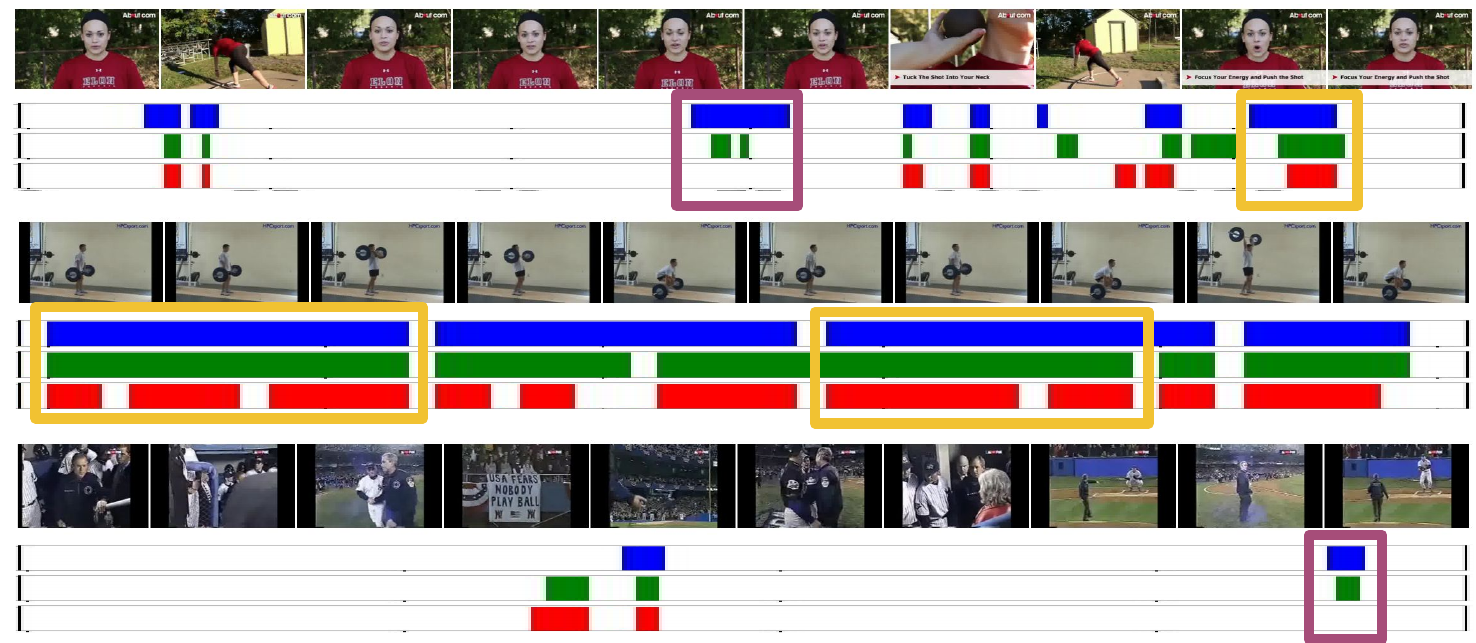}
    \caption{\textbf{Qualitative Comparison:} The ground truth is in blue, our detections are in green, and a baseline without a graph (`FC-CASL') results are in red. The video frames are sampled uniformly across the video length. By using similarity across time segments to make our predictions, our method is able to localize larger extents of actions (yellow) and is able to develop a more general model of action classes that allows it to localize to more instances of an action (magenta).}
    \label{fig_comparison}
\end{figure*}
Our last loss is a modification of the Co-Activity Similarity Loss (CASL)~\cite{Paul_2018_ECCV}. It supervises the intermediate feature representation corresponding to video segments by both increasing the distance between foreground and background features, and decreasing the distance between foreground features of the same class.  

The foreground and background representations are the sum of time segments' intermediate feature representations weighed by their predicted classification confidence. Specifically, for a given video, let $\mb{F}_t$ represent the intermediate feature representation of time segment $t$, let $p_{i,t}$ represent the classification confidence of time segment $t$ belonging to class $i$, and let $\hat{p}_{i,t}$ represent $p_{i,t}$ after softmax normalization across all classes to segment $t$. The foreground feature representation $\mb{f}_i$, and background feature representation $\mb{b}_i$, are then calculated as:
\begin{align*}
    \mb{f}_i = \sum_{t=1}^{l} \hat{p}_{i,t}\, \mb{F}_t, \;\;\;
    \mb{b}_i = \sum_{t=1}^{l} (1-\hat{p}_{i,t})\, \mb{F}_t
\end{align*}
where $l$ is the total number of time segments in the video.

For a video $j$ and its ground truth action class $i$, the foreground $\mb{f}_i^j$ and background $\mb{b}_i^j$ feature representations are obtained. For any two videos $j$ and $k$ belonging to the same class $i$, their foreground and background representations can then be used to calculate the Co-Activity Similarity Loss:
\begin{align*}
    L_{CASL}^{j,k,i} = &\max(0,  \bar{f}(\mb{f}_i^j,\mb{f}_i^k) - \bar{f}(\mb{b}_i^j,\mb{f}_i^k) + 0.5) \\ + &\max(0, \bar{f}(\mb{f}_i^j,\mb{f}_i^k) - \bar{f}(\mb{b}_i^k,\mb{f}_i^j) + 0.5)
\end{align*}
where $\bar{f}(a,b)$ is cosine distance and $0.5$ is the margin.

CASL was originally designed to supervise the intermediate feature representation that is used to make the final class wise predictions; i.e., an unmodified use of CASL would be on the output of our graph convolution layer. Here, we instead apply the loss on the output of $\phi$. That is, we use CASL to encourage the edge weight between two foreground segments $a$ and $b$ of the same class to be high (and the edge weight of a foreground and background segment to be low). This affects how rows of $\mb{X}$ are averaged. It does not directly supervise the learned weight matrix $\mb{W}$; $\mb{W}$ is still free to transform rows $a$ and $b$ of $\mb{G}\mb{X}$ differently. In this sense, our modified CASL (MCASL), i.e. applying CASL on $\phi$, is a less rigid imposition of the loss, one that would not be possible in a regular fully connected layer. In Section \ref{sec_exp_casl}, we show that this choice is more effective in reducing overfitting than directly supervising the intermediate feature representation.

\subsubsection{Final loss}
The final loss used to supervise the training is:
\begin{equation*}
    L_{Total} = \lambda_{1}L_{MIL}+\lambda_{2}L_{L1}+\lambda_{3}L_{CASL}.
\end{equation*}
We set $\lambda_{1} = \lambda_{2} = \lambda_{3} = 1$. These hyperparameters are set so that no one loss dominates training.

\subsection{Action classification and localization}

During test time, we input a single video, and obtain an $l\times c$ volume output $\mb{Y}$. We average the top $k$ segments per class to obtain a video-level classification prediction.

In order to obtain hard localization predictions (video segment classifications), we threshold the confidence values to ignore the lowest $5\%$ range of predictions. We merge temporally consecutive time segments that are classified as the same action into a single detection, and assign it the maximum confidence of its merged segments. We use these detections for the final evaluation. 

\vspace{-0.05in}
\section{Experiments} \label{sec_exp}
We evaluate our approach against state-of-the-art weakly-supervised temporal action localization methods.  We also analyze the effects of edge sparsity and our different losses. Lastly, we present qualitative and quantitative results that highlight the advantage of our graph-based approach over traditional methods that do not explicitly model the relationship between time segments.

\vspace{-10pt}
\paragraph{Datasets}
We present results on three datasets, of which THUMOS’14 and ActivityNet 1.2 have been previously used to evaluate weakly supervised action localization.

\vspace{-10pt}
\paragraph{\textit{THUMOS’14} \cite{idrees2017thumos}} has temporal annotations for 20 classes, with 200/211 untrimmed validation/test videos. Each video contains one or more of the 20 classes, with an average of 1.12 classes per video. We use the validation dataset for training, and the testing data for testing.
\vspace{-10pt}
\paragraph{\textit{ActivityNet 1.2} \cite{caba2015activitynet}} comprises 4819 training videos, 2383 validation videos, and 2480 test videos with withheld labels. There are a 100 action classes with an average of 1.5 temporal activity segments per video. We use the training videos as training data, and validation videos as test data. 
\paragraph{\textit{Charades} \cite{sigurdsson2016hollywood}} is composed of 9848 videos, with 7985 as training videos, and 1863 as validation videos. The videos have an average length of just 30 seconds, and feature fine grained actions such as `Putting Clothes Somewhere' and `Throwing Clothes Somewhere' performed in visually similar indoor settings. Videos have an average of 6.75 actions. We use features extracted from i3D network finetuned on Charades \cite{carreira2017quo}.

\vspace{-10pt}
\paragraph{Implementation details}

\begin{figure}[t!]
  \begin{minipage}{0.6\linewidth}
  \renewcommand{\tabcolsep}{0.1mm}
    \centering
   	\resizebox{\linewidth}{!}{
    \begin{tabular}{ccccc}
		\hline
		\multirow{2}{1pc}{Method} & {} & {mAP@IoU} & {} & {}\\
		& {0.5} & {0.7} & {0.9}\\
		\hline
		{UntrimmedNets \cite{wang2017untrimmednets}} & {7.4} & {3.9} & {1.2}\\
		{Auto-Loc \cite{Shou_2018_ECCV}} & {27.3} & {17.5} & {6.8}\\
		{W-TALC \cite{Paul_2018_ECCV}} & \textbf{37.0} & {14.6} & {-}\\
		 \hline
		{Ours} & 29.4 & \textbf{{17.5}} & \textbf{{7.5}}\\
		\hline
	\end{tabular}
}
  \end{minipage}~
  \begin{minipage}{0.4\linewidth}
  \renewcommand{\tabcolsep}{0.1mm}
    \centering
       	\resizebox{\linewidth}{!}{
    \begin{tabular}{cc}
		\hline
		{Method} & {mAP}\\
		\hline
		Sigurdsson et al. \cite{sigurdsson2016hollywood} & {12.8}\\
		SSN \cite{zhao2017temporal} & 16.4\\
		Super Events \cite{piergiovanni2018learning} & 19.4\\
		TGM \cite{piergiovanni2019temporal} & \textbf{22.3}\\
		 \hline
		{Ours} & \textbf{15.8}\\
		\hline
	\end{tabular}
}
\end{minipage}
\vspace{0.5em}
\caption{\textbf{(Left)} Localization performance on ActivityNet 1.2 val set. \textbf{(Right)} Localization performance on Charades. All methods except `Ours' are strongly supervised}
\label{table_anet}
\end{figure}



The output of $\phi$ as well as our graph layer is 1024. The output of the graph layer is passed through a $\mathrm{ReLU}$ non-linearity and then $\mathrm{L2}$ normalized before being passed to the linear classification layer. We use Dropout at $0.5$ between the graph and linear layer. The output of the classification layer is passed through a $\mathrm{Tanh}$ layer to obtain the final class confidence values. The final $\mathrm{Tanh}$ non-linearity limits the range of class confidence scores so that a standard threshold of $-0.9$ can be applied across all datasets. Using a standard threshold ensures that we do not trivially inflate performance for datasets with longer actions by predicting the full duration of each video.

Though not encountered in our experiments, the graph layer’s matrix multiplication $\mb{GX}$ can run into GPU memory limitations for large graphs. During train time, the number of time segments per graph can be limited, and  during test time $\mb{G}$ and $\mb{GX}$ can be calculated offline on CPU, or in smaller row wise chunks on GPU as a solution.

We train for 250 epochs with Adam \cite{kingma2014adam} at a learning rate of $0.001$. During both training and testing we build $\mb{G}$ from time segments from a single video at a time.

For THUMOS’14, we use a batch size of 32 videos and calculate the CAS loss for every pair of videos with the same ground truth class label. For the larger ActivityNet 1.2 and Charades, we use a batch size of 256. Since calculating the CAS loss for every pair of videos for this larger batch size increases the required training time exponentially, we fix half of each batch with video pairs that have a randomly picked class in common. The CAS loss is then only calculated for the paired videos.

\subsection{Comparison to state-of-the-art}
\begin{table}[t]
\resizebox{\linewidth}{!}{
	\centering	
	\begin{tabular}{ccccccc}
	\hline
		{} & {} & {} & {mAP@IoU} & {} & {} & {}\\
		{Method} & {0.1} & {0.2} & {0.3} & {0.4} & {0.5} & {Cls}\\
		\hline 
		{HAS \cite{singh2017hide}} & {36.4} & {27.8} & {19.5} & {12.7} & {6.8} & {-} \\
		{UntrimmedNets \cite{wang2017untrimmednets}} & {44.4} & {37.7} & {28.2} & {21.1} & {13.7} & {74.2}\\
		{STPN (UNTF) \cite{nguyen2017weakly}} & {45.3} & {38.8} & {31.1} & {23.5} & {16.2} & {-}\\
		{STPN (I3DF) \cite{nguyen2017weakly}} & {52.0} & {44.7} & {35.5} & {25.8} & {16.9} & {-}\\
		{AutoLoc \cite{Shou_2018_ECCV}} & {-} & {-} & {35.8} & {29.0} & {21.2} & {-}\\
		{W-TALC (UNTF) \cite{Paul_2018_ECCV}} & {49.0} & {42.8} & {32.0} & {26.0} & {18.8} & {-} \\
		{W-TALC (I3DF) \cite{Paul_2018_ECCV}} & {55.2} & {49.6} & {40.1} & {31.1} & {22.8} & {85.6} \\
		{MAAN \cite{yuan2018marginalized}} & {59.8} & {50.8} & {41.1} & {30.6} & {20.3} & {94.1}\\
		{Ours } & \textbf{63.7} & \textbf{56.9} & \textbf{47.3} & {\textbf{36.4}} & {\textbf{26.1}} & {\textbf{94.2}}\\
		\hline
		{STAR* \cite{xu2018segregated}} & {\textbf{68.8}} & {\textbf{60.0}} & {\textbf{48.7}} & {34.7} & {23.0} & {-}\\
		{Ours } & {63.7} & {56.9} & {47.3} & {\textbf{36.4}} & {\textbf{26.1}} & {\textbf{94.2}}\\
		\hline
	\end{tabular}
}
\caption{Localization performance on Thumos'14 test set. The last column shows video classification performance. Asterisk indicates the method uses additional annotation.}
\label{table_thumos}
\end{table}


Table~\ref{table_thumos} and Figure~\ref{table_anet} (left) show weakly-supervised temporal action localization results on THUMOS’14 and Activity 1.2, respectively. We use mean average precision (mAP) to calculate localization accuracy at different overlap thresholds. Overlap threshold is used to determine the minimum required overlap between a ground truth occurrence and a prediction for it to count as a true positive.

For THUMOS’14, our method outperforms all previous methods at the challenging overlap threshold of $0.5$, with a margin of more than $3$ mAP points. This gap in performance is retained even when comparing against STAR~\cite{xu2018segregated} which uses additional annotation in the form of the number of times an action occurs in a video during training. Similarly, we outperform previous methods on ActivityNet 1.2 at higher overlap thresholds. To demonstrate localization ability independent of classification, we also calculate mAP for ground truth action classes. This results in $19.7\%$ and $8.2\%$ mAP at 0.7 and 0.9 IoU for ActivityNet, and a slight increase at 0.5 IoU to $63.9\%$ for THUMOS’14. 

Figure~\ref{table_anet} (right) shows additional results of our method on Charades. While our method is 6.5 points below the state-of-the-art in a fully supervised setting, it is 3 points higher than its original fully supervised baseline and presents a challenging weakly supervised baseline for future methods to compare with. Like previous methods, we report mAP for 25 equally spaced time points in each video.

\subsection{Ablation studies}
We next study the effect of our three losses. In particular, we study the effect of CASL by showing that it is more effective with a graph-based method than an approach that does not explicitly cluster time segments together. We show that the modified CASL is able to do better by guarding against over-fitting. Last, we inspect how $k$ should be set for the top $k$ multi instance learning loss.

\subsubsection{Graph supervision}

\begin{table}[t]
	\centering	
	\footnotesize
	\begin{tabular}{cccccc}
		\hline
		{} & {} & {mAP@IoU} &{} &  {} & {}\\
		{Method} & {0.1} & {0.2} & {0.3} & {0.4} & {0.5}\\
		\hline
		{Baseline} & {26.1} & {19.4} & {13.1} & {8.9} & {5.8}\\
        {MCASL} & {26.7} & {20.8} & {14.7} & {9.9} & {6.2}\\
        {L1} & {55.3} & {46.9} & {39.0} & {28.5} & {19.6}\\
        {L1+MCASL} & \textbf{63.7} & \textbf{56.9} & \textbf{47.3} & {\textbf{36.4}} & {\textbf{26.1}}\\
		\hline
	\end{tabular}
\caption{Ablation study of different constraints on our appearance similarity graph on Thumos `14 test set.}
\label{table_ablation}
\end{table}

We first analyze the importance of each constraint on the appearance similarity graph. The appearance similarity graph uses an L1 loss to encourage non-uniform edge weights, and a co-activity similarity loss (CASL) on $\phi$ to supervise edge clustering. Table \ref{table_ablation} shows the results of our ablation study. L1 loss is most crucial for performance, as it more than triple the performance at 0.5 overlap. MCASL provides the next significant improvement: a 8.4 mAP improvement at 0.1 IoU threshold. While MCASL improves performance of the baseline model, it is most useful when working with an L1 loss. This indicates MCASL is more useful when working with a sparse graph.

\subsubsection{Modified co-activity similarity loss}\label{sec_exp_casl}
\begin{table}[t]
    \footnotesize
	\centering	
	\begin{tabular}{cccccc}
		\hline
		{} & {} & {} & {mAP@IoU} & {} & {} \\
		{Method} & {0.1} & {0.2} & {0.3} & {0.4} & {0.5}\\
		\hline
		{FC-CASL 1024}	&	{55.1}	&	{47.9}	&	{38.4}	&	{29.4}	&	{18.3}\\
		{FC-CASL 2048} 	&	{55.4}	&	{48.3}	&	{40.0}	&	{30.3}	&	{19.8}\\
		{CASL-Graph}	&	{57.7}	&	{50.9}	&	{42.0}	&	{32.1}	&	{22.5}\\
		{MCASL (Ours)} & \textbf{63.7} & \textbf{56.9} & \textbf{47.3} & {\textbf{36.4}} & {\textbf{26.1}}\\
		\hline
	\end{tabular}
\caption{Using a graph with CASL (last two rows) is more effective than using regular linear layers (FC-CASL rows) since it explicitly utilizes relationships between temporal segments.}
\label{table_casl}
\end{table}


We next analyze the effect of the co-activity similarity loss. 

We develop a baseline model that uses the CASL loss without a graph convolution layer to contrast it with our graph-based approach. Specifically, the model uses a fully-connected layer instead of a graph layer, but is otherwise identical. The resulting model `FC-CASL 1024' has a 1024 dimension intermediate output like our graph model. We also train a higher-capacity model `FC-CASL 2048' that has a larger intermediate layer with a 2048 dimension output, which is roughly the same number of learnable parameters as ours. These baseline models are very similar to the model in \cite{Paul_2018_ECCV}, except they have the same non-linearities as our network. These are also equivalent to our network without a learned similarity metric $\phi$, but a fixed identity adjacency matrix $\mb{G}$. We additionally develop a baseline model that uses the original CAS loss `CASL-Graph': instead of applying CASL on the output of $\phi$ as done in our model, we apply it to the output of our graph layer. Thus, the only difference between this baseline and the `FC-CASL' baselines is the graph layer.

Table~\ref{table_casl} shows the results. Applying the CASL loss on the output of the graph layer `CASL-Graph', leads to a $\sim$3 mAP improvement over the `FC-CASL' baselines. This points to the superiority of using a graph based approach for weakly-supervised action localization versus relying on conventional linear layers. In addition, the better performance of our full model compared to `CASL-Graph' shows that our modified CASL which supervises input feature clustering, rather than intermediate network features is a better method for providing supervision. By tracking testing performance throughout training, we find that `CASL-Graph' begins to overfit midway through training after reaching peak performance at $59$ mAP at 0.1 IoU. On the other hand, `Ours-MCASL' reaches higher peak performance and then maintains it through the end of training since it is not modifying the actual intermediate feature representation of the network, but only modifies how the input I3D features are clustered for further inference.
\subsubsection{MIL Loss Parameter}\label{sec:denosize}




\begin{table}[t]
\begin{center}
\resizebox{\linewidth}{!}{
	\centering	
	\renewcommand{\tabcolsep}{0.1mm}
	\begin{tabular}{cc|cc|cc|cc}
		\hline
		\multirow{3}{1pc}{$d$} &{}& \multicolumn{2}{c|}{THUMOS} & \multicolumn{2}{c|}{ActivityNet} & \multicolumn{2}{c}{Charades} \\
		 &{Video}& {mAP} & {Test Data}& {mAP} & {Test Data}& {mAP} & {Test Data}\\
		   &{\%} & {@ 0.5 IoU} & {\%}& {@ 0.5 IoU} & {\%}& {Per Frame} & {\%}\\
		\hline
		{1 }&{50-100\%}   & {18.5}  & {2.8}  & \textbf{29.4}  & {57.2}  & {14.9}  & {76.9}\\
		{2 }&{25-50\%}   & {44.9}  & {3.8}  & {5.5}  & {19.0}  & {15.4}  & {82.0}\\
		{4 }&{12.5-25\%}   & {58.4}  & {14.1}  & {1.7}  & {14.4}  & {15.2}  & {75.5}\\
		{8 }&{0-12.5\%}   & \textbf{63.7}  & {93.9}  & {1.4}  & {18.8}  & {13.8}  & {15.4}\\
		{Random}   &{}& {39.0} & {-} & {14.3}  & {-}  & \textbf{15.8}  & {-}\\
		\hline
	\end{tabular}
	}
	
\caption{Setting hyperparameter $d$ to correspond with expected action duration results in the best performance across datasets.}
\vspace{-1em}
\label{table_deno}
\end{center}
\end{table}
\begin{figure*}[t]
    \centering
    \includegraphics[width=0.9\textwidth]{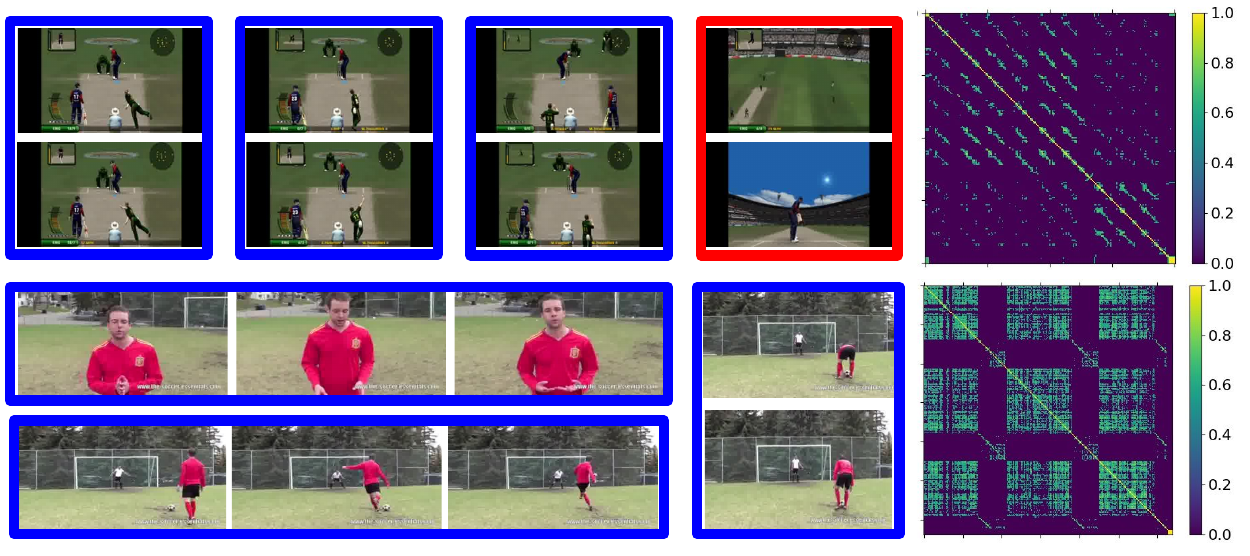}
    \caption{\textbf{Visualizing graphs:} Strongly connected graph cliques are shown in blue. In red we show segments that are considered very dissimilar to the foreground segments. The corresponding adjacency matrix for each example is shown on the right.}
    \label{fig_graph_viz}
\end{figure*}
As explained in Section~\ref{sec_mil}, the multi instance learning loss is calculated over the average of the top $k$ predictions of each class. $k$ is chosen to be $1/8$ of the length of a video by setting parameter $d$. While $d=8$ works well for THUMOS’14, it is not optimal for ActivityNet and Charades. 

Generally speaking a smaller $d$ (or larger $k$) results in longer detections as the MIL loss is backpropagated to more time segments at every iteration. Table \ref{table_deno} shows the performance of our system for different values of $d$ against the percent of test videos that feature activities with corresponding duration. The $d$ that results in the best performance mimics the activity duration bias for each dataset; 57\% of ActivityNet test videos feature actions that last more than half the video length, so setting $d=1$ during training results in the best performance. With very short action durations, THUMOS’14 performs best with a large $d$ or shorter predictions. Without prior knowledge of typical activity duration, or a temporally labeled validation set that can be used to set $d$, one useful strategy is to randomly choose a value for $d$ for each training iteration. The last row shows results where $d$ is randomly selected from the set $\{1,2,4,8\}$ every training iteration. With a balanced activity duration, `Random' is the best strategy for Charades, and for both ActivityNet and THUMOS’14 results in performance that is significantly better than the worst $d$ setting, but about half of the optimal level. Estimating $d$ without any temporal annotation is an interesting direction for future research.
\subsection{Qualitative results}

Figure \ref{fig_comparison} shows some qualitative results. Ground truth, our results, and the `FC-CASL' baseline results are shown in blue, green and red, respectively, for videos from different classes. Using a graph allows our network to localize actions with more overlap (in yellow). This is most apparent in the second row, where our detections are not split up and wider than the baseline's. Our model is also able to localize more occurrences of different actions; in magenta we show instances that are not detected by `FC-CASL' but are detected by our method. We include more qualitative examples and failure cases in the supplementary materials.

\subsection{Visualizing Graphs}
In Figure \ref{fig_graph_viz} we show the adjacency matrix of two graphs, and the nodes that form high edge cliques in these graphs. All images are \emph{not} temporally neighboring, and taken from different points in the video. Graph cliques are surrounded in a blue box. Segments that are considered dissimilar to the foreground segments are surrounded in a red box. 

In the cricket bowling video, the graph forms cliques from parts of cricket bowling so that the start of the ball throw forms one cluster, the arm swing forms another, and so on. The segments considered least similar to bowling segments are shown in red and show batting, and a zoomed out view of the stadium; segments with very little relevance to the bowling action. 

The second example shows a video with three distinct cliques. The video features a man explaining how to score a soccer penalty, and then demonstrating it repeatedly. The largest clique lumps together nodes where the man is facing the camera and talking. Another clique comprises the action right before the soccer penalty -- placing the ball and taking the starting position. The last clique lumps together the actual soccer penalty.

These examples show some interesting ways the graph can cluster nodes -- it can cluster together subactions of an action class, and structured activities that may be relevant to, but distinct from, the action class.

\section{Conclusion}
We presented a novel approach for weakly supervised temporal action localization. Without frame level annotation during training, an action localization system must necessarily infer action categories from the similarity and difference between time segments of videos. Despite this, current methods do not make explicit use of appearance and motion similarity between time segments to inform predictions. In contrast, our method makes explicit use of similarity relationships between time segments by using graph convolutions. As a result, it is able to harness similarity relationships to develop a better model of each action category, and is consequently able to localize actions to a fuller extent. We pushed the state of the art on Thumos'14 and ActivityNet 1.2 for weakly supervised action localization, and presented the first results on Charades. We demonstrated quantitatively and qualitatively that a baseline approach that does not use graph similarity achieves inferior performance. Last, we demonstrated through ablation studies the importance of each component of our system, and presented analysis of the weaknesses of our approach.

\vspace{-10pt}
\paragraph{Acknowledgements.} This work was supported in part by a grant from Swedish Research Council Formas, Hellman Fellowship, and ARO YIP W911NF17-1-0410.

{\small
\bibliographystyle{ieee}
\bibliography{main}
}

\end{document}